\documentclass[letterpaper,10pt,conference]{ieeeconf}
\IEEEoverridecommandlockouts

\usepackage[OT1]{fontenc}
\usepackage{cite}
\usepackage{amsmath,amssymb,amsfonts}
\usepackage{graphicx}
\usepackage{textcomp}
\usepackage{xcolor}
\usepackage{url}
\usepackage{hyperref}
\usepackage{array}
\usepackage{tabularx}
\usepackage{booktabs}

\hypersetup{
    colorlinks=true,
    linkcolor=black,
    citecolor=black,
    urlcolor=black,
    pdftitle={AVALON: Zero-Shot Indoor Navigation with Human-Seeded Topological Memory},
    pdfauthor={},
    pdfsubject={},
    pdfkeywords={}
}

\title{\LARGE\bfseries AVALON: Zero-Shot Indoor Navigation with Human-Seeded Topological Memory}

\author{
Jing Xia$^{1}$, Changyan Li$^{1}$, Yuduo Wang$^{1}$, Guiyin Tian$^{1}$, Bo Wang$^{2}$, Yifan Liao$^{1}$, Mingjun Li$^{1}$, Shiyan Hu$^{1}$%
\thanks{$^{1}$Department of Data and Systems Engineering, The University of Hong Kong, Hong Kong SAR, China.}%
\thanks{$^{2}$Department of Electrical and Computer Engineering, The University of Hong Kong, Hong Kong SAR, China.}%
}

\begin{document}

\maketitle
\thispagestyle{empty}
\pagestyle{empty}

\begin{abstract}
Indoor service robots often enter unmapped buildings whose layouts and task constraints are already familiar to users. We address this \emph{human-familiar, robot-unseen} setting with \textbf{AVALON}, a topological navigation framework organized around human-seeded layered spatial memory, node-verified topological execution, and task-triggered, evidence-gated human--robot memory co-updating.
The robot has no prior on-site traversal or site-specific training. Deployment requires only a floor plan, place clues, and an initial topology-odometry alignment.
Visual depth point-goal navigation leverages verified node arrivals to perform translational re-anchoring, without requiring a robot-constructed global SLAM map. Human--robot memory co-updating fuses complementary sources of information. Human operators specify task constraints that are unobservable via vision and report offline environmental modifications, whereas the robot provides physical-state evidence through active in-situ validation. Task relevance governs whether human-reported environmental changes undergo robot validation during the active mission. Field-collected perceptual evidence gates memory updates, maintaining an explicit separation between human-specified permissions and robot-observed traversability. Our experiments on a Unitree Go2 show that AVALON with X-NavDP completes a 12.81~m Intersection-to-Dining mission in 46.4~s ($5/5$) with no mapping, 40.9\% less robot time than DimOS-Go2 (78.5~s, $4/5$), while policy-only X-NavDP times out in all of the 5 runs. These demonstrate the superiority of the proposed topological navigation framework. 
\end{abstract}

\makeatletter
\global\@topnum\z@
\makeatother
\begin{figure*}[t]
\centering
\includegraphics[width=5.8in]{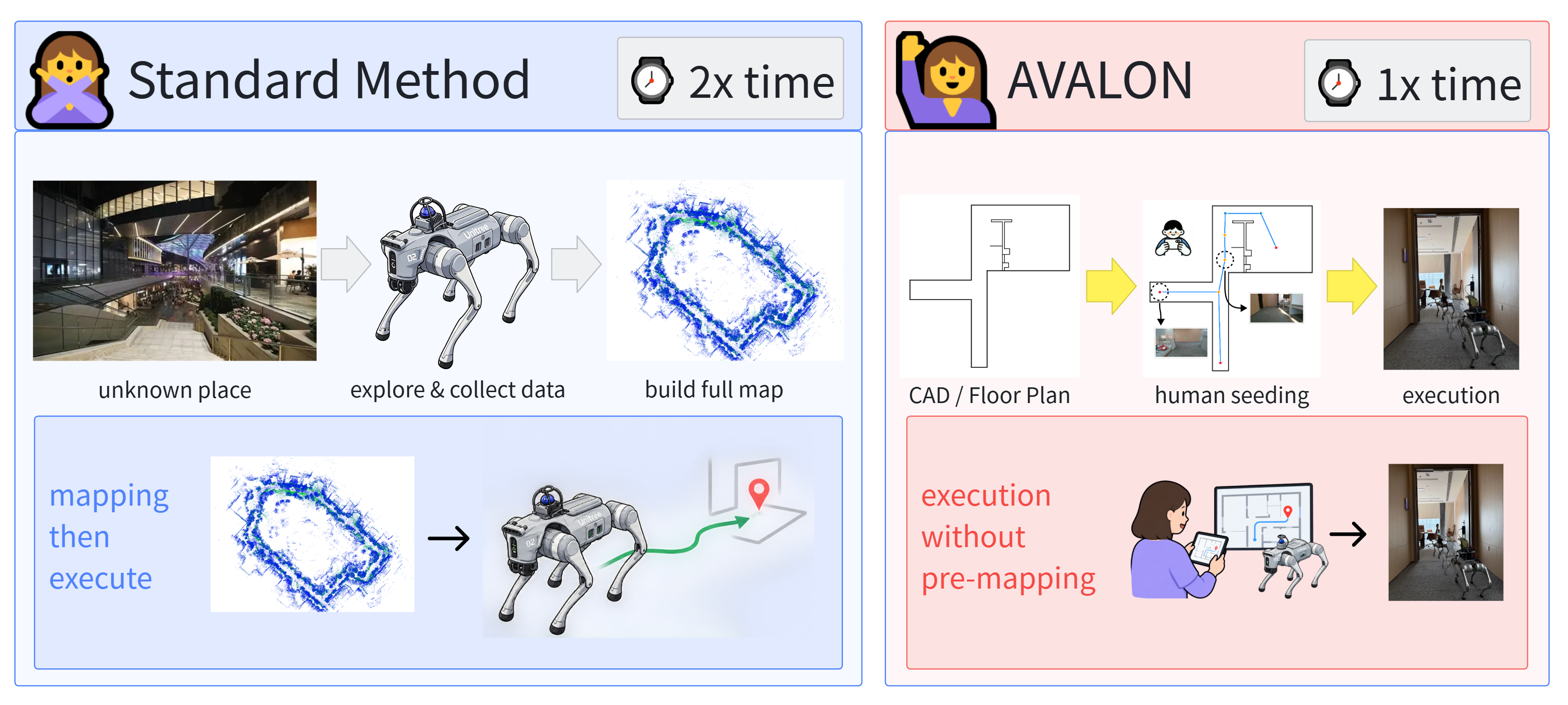}
\caption{Standard mapping-then-execute versus AVALON: a user seeds a floor-plan graph and the robot executes the first mission without a mapping pass. Measured time is in Table~\ref{tab:main-results}.}
\label{fig:teaser}
\end{figure*}


\section{Introduction}
\label{sec:introduction}

\begin{table*}[t]
\caption{Comparison on how first-mission spatial memory is obtained and used, where memory co-updating denotes whether user and the robot write back to the same node/edge memory in the first mission.}
\label{tab:contrast}
\centering
\scriptsize
\setlength{\tabcolsep}{3pt}
\renewcommand{\arraystretch}{1.15}
\begin{tabularx}{\textwidth}{@{}l >{\centering\arraybackslash}X >{\centering\arraybackslash}X >{\centering\arraybackslash}X >{\centering\arraybackslash}X@{}}
\toprule
Method & Spatial prior & Robot mapping & Localization & Memory co-updating \\
\midrule
FloNa & floor plan & no & metric pose & no \\
osmAG-Nav & CAD & no & LiDAR pose & no \\
RTAB-Map & none & dedicated mapping & metric map & no \\
NTS & none & exploration & node & no \\
\textbf{AVALON} & \textbf{CAD + user clues} & \textbf{no (seed)} & \textbf{node verify} & \textbf{yes} \\
\bottomrule
\end{tabularx}
\end{table*}

Indoor service robots often enter unmapped buildings whose layouts, routes, and closures are already familiar to users. This \emph{human-familiar, robot-unseen} setting offers a remarkable opportunity for rapid deployment of robots. Users can supply a floor plan and delivery instructions before the first mission (Fig.~\ref{fig:teaser}). The robot must turn these priors into executable navigation memory and detect disagreements with observations.

Dedicated pre-mapping and online topological exploration~\cite{chaplot2020neural,kim2023tsgm,chen2026hgr} both require robot data collection that may delay service. Existing CAD, BIM~\cite{vega2023bimslam}, and user knowledge offer alternative spatial priors, but these must be grounded in observations and updated as conditions change.

In this paper, we propose \textbf{AVALON} (Asymmetric Visual-Architectural Layered Online Navigation), built around \emph{seed, execute, and co-update}. 
%
A spatial memory framework is constructed based on floor-plan topological skeletons with visibility-aware constrained traversal hops, and paired with editable task memory that shares unified node and edge identifiers. Multi-modal contextual clues, including casual imagery, linguistic descriptions, and RoadBook annotations, are explicitly associated with these topological primitives. The topological skeleton provides global route planning priors for the robot, while the attached multi-modal clues serve as verification signatures to validate the arrival at target topological nodes. Local point-goal control governs the robot's inter-node navigation, and node arrival is rigorously validated via fused visual-semantic observations and odometry measurements.

Successful arrival confirmation enables translational re-anchoring and progressive route execution. In case of visual observation mismatch with the expected node template, the robot adaptively refines its localization hypothesis. For object search tasks, real-time perceptual inference is deployed to verify object presence within the corresponding spatial regions, and the collected perceptual evidence is archived to support subsequent downstream missions. Robot onboard observations continuously record environmental physical states, whereas human operators supplement task-specific constraints and update unobserved environmental variations. Benefiting from the initialized prior memory seed, the system can execute initial missions without standalone dedicated mapping procedures. Furthermore, human-robot collaborative memory updating maintains the timeliness and accuracy of the memory bank, accommodating dynamic changes in workspace layouts and task specifications.

AVALON replaces globally precise localization with human-seeded, task-sufficient spatial grounding. Topology specifies where to go, sparse semantic anchors re-ground route progress, and terminal perception confirms what the task seeks. AVALON turns human-held spatial knowledge into executable memory for a robot's first mission by coupling a human-seeded structural prior, node-verified local execution, and task-triggered evidence acquisition through stable graph IDs. The main contributions of this paper can be summarized as follows.
\begin{enumerate}
    \item \textbf{Human-Seeded Layered Spatial Memory.} We propose to decouple a floor-plan skeleton from editable task memory, and use shared node-edge identifiers to associate human-provided clues, task constraints, and robot sensory evidence. This design eliminates the requirement for a dedicated pre-mapping traversal prior to deployment.
    \item \textbf{Node-Verified Topological Execution.} The robot navigates along human-seeded routes via local graph transitions. Node-associated visual and textual clues validate arrival and re-anchor execution progress. Mismatches between real observations and expected node signatures trigger incremental revision of the localization hypothesis.
    \item \textbf{Task-Triggered, Evidence-Gated Human-Robot Memory Co-Updating.} Graph identifiers are shared across execution and memory maintenance modules. Human permissions remain authoritative, while change notifications invoke on-site physical validation only when relevant to the active task. Physical world states are updated exclusively using spatially localized robot evidence, and verified execution outcomes refresh the spatial memory graph to support subsequent missions. 
    \item \textbf{Experimental Validation.} We validate our proposed closed-loop pipeline on a Dining-to-Meeting-Room navigation scenario involving a human-reported cart at the entryway. On a Unitree Go2, AVALON with X-NavDP completes a 12.81~m Intersection-to-Dining mission in 46.4~s ($5/5$) with no mapping, 40.9\% less robot time than DimOS-Go2~\cite{dimensionalos2026dimos} (78.5~s, $4/5$), while policy-only X-NavDP times out in all of the 5 runs. In the five AVALON + X-NavDP Dining-to-Meeting-Room trials reported in Table~\ref{tab:main-results}, user reports of environmental changes trigger on-site robot verification before the entrance node condition is updated. AVALON successfully completes the report--verification--update sequence and subsequently reaches the meeting room in all five trials. These demonstrate the superiority of the proposed topological navigation framework.
\end{enumerate}

\section{Related Work}
\label{sec:related_work}

Table~\ref{tab:contrast} compares the first-mission memory in terms of the categories including what prior exists, whether the robot must map, how it localizes, and whether user later edit the same memory with the robot.

\subsection{Floor Plans and CAD Priors}
Floor plans have been used to localize a camera~\cite{howard2021lalaloc,howard2022lalalocpp} and to condition a learned navigation policy~\cite{li2025flona}. Imperfect drawings can also guide exploration~\cite{luperto2019robot}. Other user-provided sketches include posted signs~\cite{zimmerman2025signloc,zimmerman2026signscene} and hand-drawn maps~\cite{tan2025hamnav}. CAD and BIM can initialize a workspace model before any robot visit~\cite{vega2023bimslam,zhang2025cad_osmag}.

osmAG-Nav demonstrates how CAD priors can support planning on the passage graph together with structure-aware LiDAR localization~\cite{zhang2026osmag_nav}. In contrast, AVALON addresses a different first-deployment question, i.e., how architectural priors and user-held task knowledge can become executable shared memory before the robot traverses the site. It converts a floor plan into a visibility-constrained point-goal skeleton, while stable node and edge IDs bind user photos, language descriptions, and RoadBook notes to that structure. During execution, these clues verify semantic progress and gate translational re-anchoring. The same IDs subsequently organize robot evidence and user edits. AVALON therefore couples first-mission routing, semantic grounding, and human-robot memory co-updating without a dedicated robot mapping pass or LiDAR-map localization.

\subsection{Two-Stage Mapping and Exploration}
A common first-mission pattern is two-stage, i.e., the robot maps or explores, then executes. Metric SLAM such as RTAB-Map builds a global map in a dedicated pass and later localizes against it~\cite{labbe2019rtabmap}. Topological methods such as NTS, TSGM, and HGR grow a node-edge graph while exploring~\cite{chaplot2020neural,kim2023tsgm,chen2026hgr}. SSMG-Nav and MG-Nav likewise construct semantic or geometric memories online~\cite{niu2026ssmgnav,wang2025mgnav}. OPENNAVMAP fuses multi-session image trajectories into a shared graph with full-pose estimates~\cite{jiao2026opennavmap}. In all of these cases the robot must traverse the site before it has an executable map. In contrast, AVALON seeds the backbone from the floor plan so the first mission already has nodes and edges. Verified arrivals re-anchor translation at those nodes.

\subsection{Map-Free Policies}
Goal-directed policies can run with no site map~\cite{shah2023gnm,shah2023vint,sridhar2023nomad,cai2025navdp,yang2026xnavdp}, and zero-shot semantic navigation seeks objects in unseen scenes~\cite{yokoyama2024vlfm,zhang2024navid,yin2025unigoal}. AVALON interfaces with NavDP and X-NavDP through robot-frame point goals and evaluates both policies without its graph. These policies receive no user floor plan and are not treated as object-goal benchmarks. osmAG-LLM uses map-guided object confirmation, while Bee-Nav separates transit from terminal visual guidance~\cite{xie2026osmag_llm,ou2026beenav}. AVALON instead reaches a graph-associated candidate region and confirms the requested object from current observations. Timed trials are room-to-room with terminal confirmation, not open-ended search. Here, \emph{zero-shot} means no on-site mapping or site-specific training, given only a human-provided floor plan and initial topology-odometry alignment.

\subsection{Human-Robot Memory Updates}
Human-in-the-loop semantic SLAM lets operators correct geometry and labels on a robot-built map~\cite{ribeiro2025hicsslam}. In contrast, AVALON's co-updating is at task level. Human users encode navigational permissions and environmental updates that are inherently unobservable via onboard vision, while the robot grounds memory entries with physical occupancy measurements acquired in situ. Human-reported modifications relevant to the active task undergo on-site robotic validation before physical-state memory is rewritten; other reports remain pending until a later relevant task. We validate this closed-loop human--robot memory update pipeline in real-world Dining-to-Meeting-Room navigation deployments.

\section{Method}
\label{sec:method}

\subsection{Problem Formulation and System Overview}
\label{sec:overview}
Assuming that a CAD or floor-plan drawing is given, we build a node-edge skeleton $G=(V,E)$ with stable IDs from it. Place photos, language descriptions, and RoadBook notes are then attached to the nodes on this graph. A single alignment between the topology frame and robot odometry is required at initialization. The robot has no on-site traversal, site-specific training, or robot-built global SLAM map. A mission specifies a room, an object description, or a reference image~\cite{yin2025unigoal}, which the memory associates with one or more graph nodes. For room navigation, success is declared when the destination node passes arrival verification. For object search, verified arrival at a candidate node determines where to search, but success is declared only after live perception detects the requested object.

\begin{figure*}[!t]
\centering
\includegraphics[width=\textwidth]{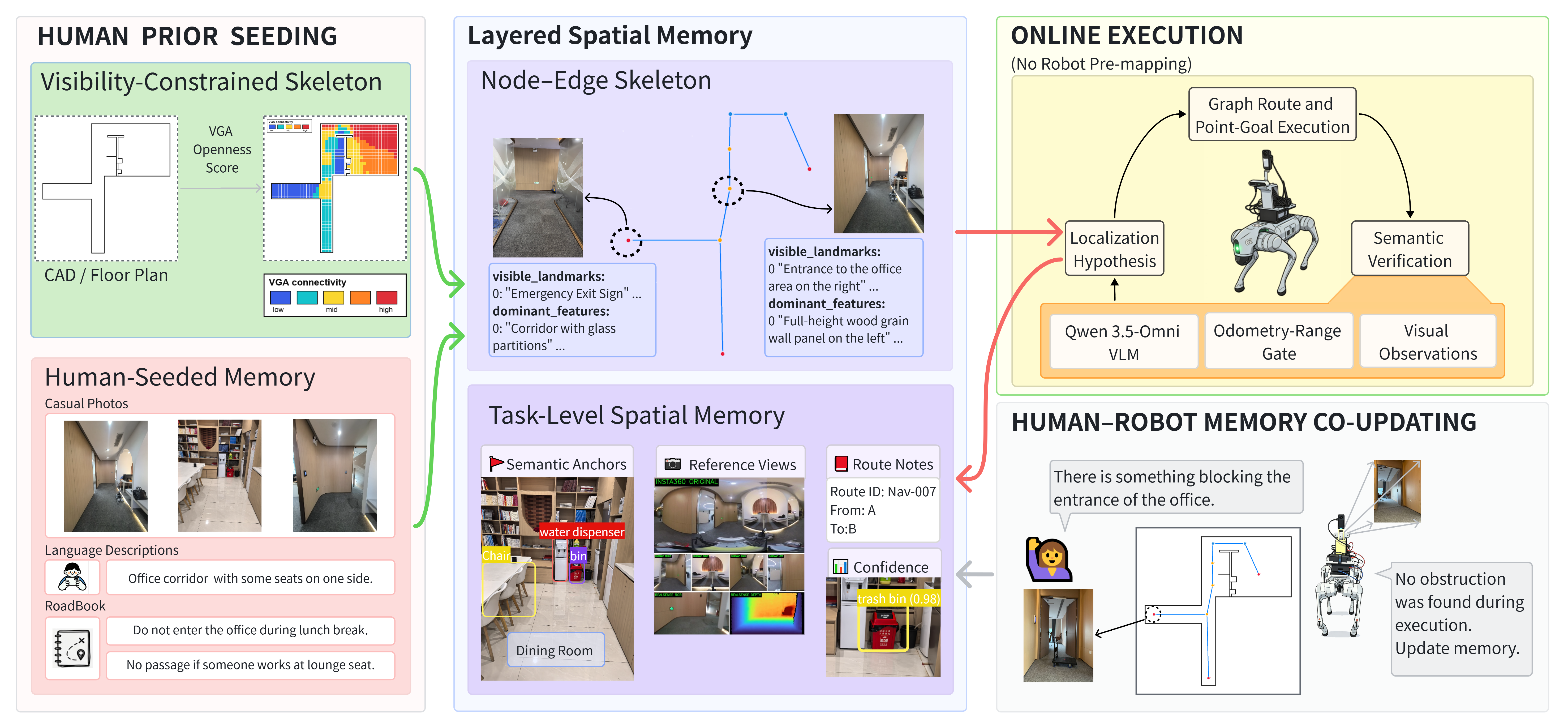}
\caption{AVALON system overview. Users seed a stable structural skeleton from CAD/floor plans and attach task-level spatial memory (casual photos, text, and RoadBook notes) on stable IDs. In actual operations, the robot follows the seeded graph with local motion, verifies semantic-node arrival, and re-anchors translation without pre-mapping. Verified visits write back to the same IDs. Users can edit the same memory between deployments.}
\label{fig:architecture}
\end{figure*}

\begin{figure}[!t]
\centering
\includegraphics[width=\columnwidth]{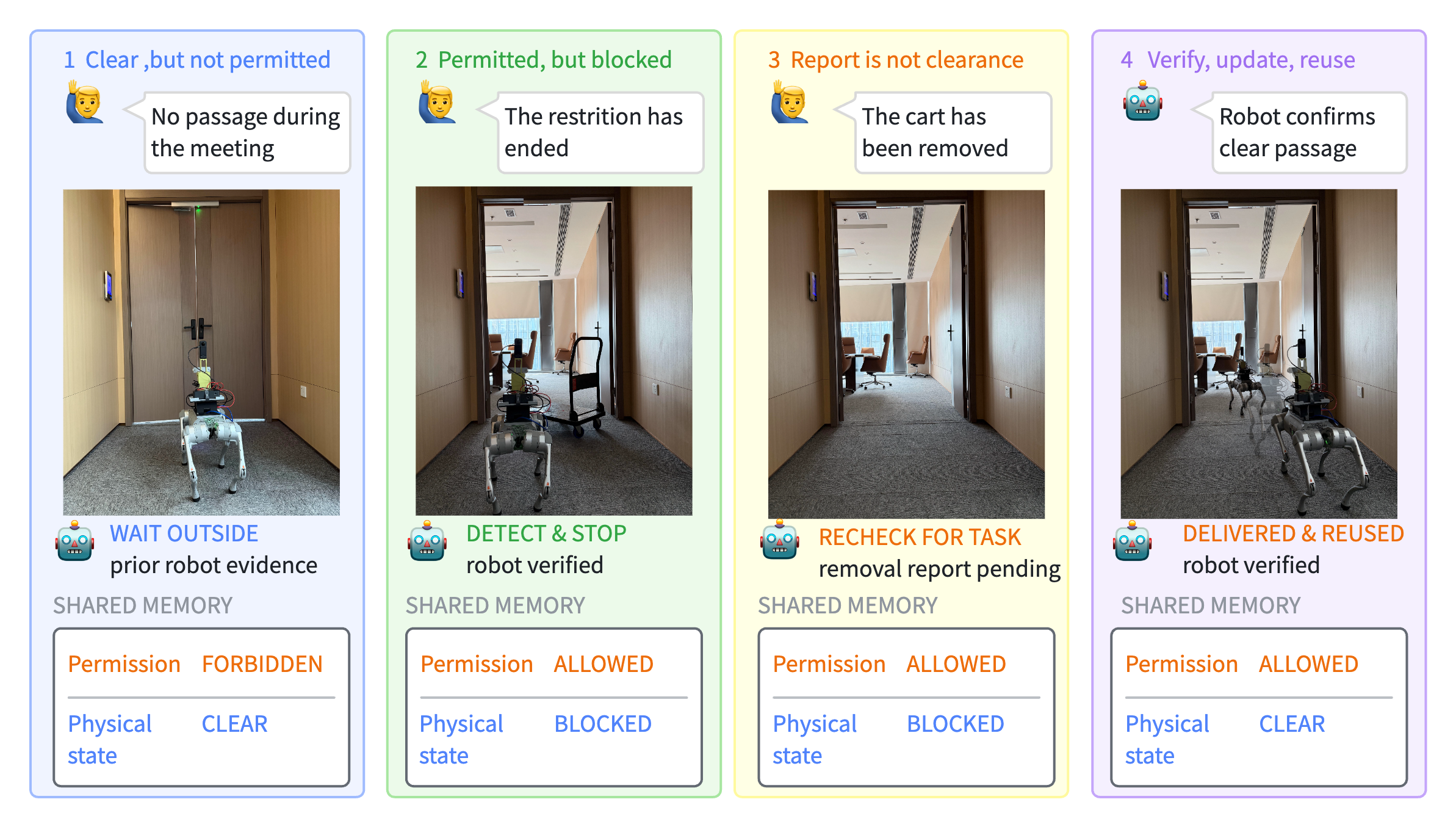}
\caption{Shared permission and physical state at one doorway. Panel titles and memory boxes state the four cases. The text gives the update rules.}
\label{fig:maintenance-mechanism}
\end{figure}

Figure~\ref{fig:architecture} outlines the three stages of user-seeded memory construction, verified execution, and evidence-gated co-updating.
Section~\ref{sec:seeded-memory} first builds the CAD-derived skeleton, then attaches human clues.
Section~\ref{sec:execution} follows the graph through a robot-frame point-goal interface, instantiated here with RGB-D local policies~\cite{cai2025navdp,yang2026xnavdp}, and re-anchors translation at verified semantic nodes.
Section~\ref{sec:co-maintenance} keeps human permissions separate from physical state and accepts physical-state updates only with localized robot evidence (Fig.~\ref{fig:maintenance-mechanism}).

\subsection{Human-Seeded Layered Spatial Memory}
\label{sec:seeded-memory}
In the construction, a CAD or floor-plan raster first produces the node-edge skeleton, and photos, language descriptions, and RoadBook notes are attached by users to the nodes in the graph. Route connectivity is therefore separate from the evidence used to recognize places.

\subsubsection{Structural backbone}
In contrast to CAD-to-osmAG's area-passage representation~\cite{zhang2025cad_osmag}, our \emph{visibility-constrained skeleton construction} produces straight local hops. Wall thresholding, exterior flood fill, and connected-component filtering extract a walkable pixel set $\mathcal F\subset\mathbb{Z}^2$ from a floor-plan raster with known resolution and origin. We skeletonize $\mathcal F$, group endpoints and junctions into vertices, and prune short or low-clearance branches. When an endpoint chord leaves $\mathcal F$, intermediate waypoints are inserted along the chain. Shortcutting removes a waypoint only if the resulting segment remains in $\mathcal F$. The vertex set $V$ comprises these endpoints, junctions, and inserted waypoints. Edges in $E$ join successive vertices on retained chains, not every mutually visible pair. Each hop is a free-space raster segment:
\begin{equation}
 (u,v)\in E \;\Longrightarrow\; \ell(q_u,q_v)\subseteq\mathcal F,
 \label{eq:nav-skeleton}
\end{equation}
where $q_u,q_v\in\mathcal F$ are node pixels and $\ell(q_u,q_v)$ is the Bresenham segment between them~\cite{black2020bresenham}. Metric positions $p_u,p_v$ follow from the same resolution and origin. Route costs can be generic such as Euclidean length $w_{uv}=\|p_u-p_v\|_2$. These geometric constraints support route generation but do not replace footprint clearance or online obstacle checks. After $E$ is formed, a grid visibility preview on $\mathcal F$ may annotate each vertex with a VGA openness score~\cite{turner2001isovists} copied from nearby free-space cells. Hop visibility in~\eqref{eq:nav-skeleton} (whether a straight edge stays in $\mathcal F$) is distinct from this per-node openness (how much free space is visible from that vertex).

\subsubsection{Node roles and memory attachment}
Users then attach photos, language descriptions, or RoadBook notes to nodes, and can change node roles and edit edges between deployments, while keeping the same IDs. The editable graph distinguishes \emph{semantic nodes} from \emph{connection waypoints}. Both store metric positions and participate in routing. Users can use the openness score as a clue when assigning these roles. Visually open junctions are preferred as recognition anchors, while low-openness samples are often left as waypoints. The score does not itself label a node or authorize re-anchoring. Semantic nodes may carry descriptions, visual clues, and reference or subsequently attached images. They can be queried for place recognition and become position-correction anchors after confirmation. Verification is gated on stored images, not on openness. Nodes without images, including connection waypoints, are used only for walking and do not trigger verification or translational re-anchoring. Geometric subdivision therefore does not imply a new recognizable place. Semantic references and later observations remain separately stored.

\subsubsection{Execution interface}
Routing uses the full graph, whereas semantic candidate selection uses only semantic nodes. Semantic adjacency traverses connection-waypoint chains and stops at the next semantic node, preserving candidate relationships without requiring every intermediate waypoint to be recognizable. RoadBook notes, attached after skeleton construction, record route preferences and non-visual task constraints on node IDs, keeping human permissions separate from observed physical conditions. Current execution uses route preferences and structured rule parsing and enforcement remain extensions. This representation supplies routes and recognition evidence to Section~\ref{sec:execution}, and shared update targets to Section~\ref{sec:co-maintenance}.

\subsection{Node-Verified Topological Execution}
\label{sec:execution}
The system develops a hierarchical execution pipeline as follows. The topological graph schedules sequential node transitions, a local navigation policy executes point-goal tracking, and semantically verified arrivals correct translational alignment. For object-search tasks, terminal perceptual validation independently confirms target presence to complete mission execution.

\subsubsection{Graph Route and Point-Goal Execution}
\label{sec:scoring}
Given a task associated with one or more nodes, we compute a route on $G$ from the current topological neighborhood to a goal-associated node. Semantic nodes and connection waypoints both participate in routing. Only semantic nodes later serve as recognition anchors. RoadBook entries may bias route costs but do not authorize unsafe motion.

Deployment initializes planar alignment between topology frame $M$ and odometry $O$. For the current destination $v$ with map position $p_v^M$, fixed initial rotation $R_0$, and latest translational correction $t_k$, the robot-frame goal is
\begin{equation}
 g_t^B=R_{OB}(t)^\top\bigl(R_0p_v^M+t_k-p_B^O(t)\bigr),
 \label{eq:local-point-goal}
\end{equation}
where $p_B^O$ and $R_{OB}$ are odometric position and body-to-odometry rotation, respectively. This interface can use a local navigation policy that consumes the robot-frame goal $g_t^B$ and onboard observations to produce local motion. We instantiate it with NavDP~\cite{cai2025navdp} and X-NavDP~\cite{yang2026xnavdp}, which map live RGB-D observations and $g_t^B$ to short-horizon waypoints. Reaching a geometric waypoint advances the route but does not verify node identity.

\subsubsection{Semantic Verification and Position-Only Re-Anchoring}
\label{sec:verification}
\label{sec:heading}
Upon entering a semantic-node verification region, the robot stops and holds position until verification is accepted or rejected. Live multi-view observations are compared with stored descriptions and images. Candidates are restricted to the last confirmed semantic node and its semantic neighbors, which are reached by traversing connection-waypoint chains. Therefore, geometrically nearby but topologically unrelated places are not treated as alternatives.

A vision-language model (VLM) compares the stopped robot's multi-view observations $\{I_t^{(j)}\}_{j=1}^{J}$ with the stored image-text reference set $\mathcal R_v$ of the active candidate.Its match decision and the odometry-range gate are
\begin{equation}
\begin{aligned}
 M_t(v)&=\mathbf 1\!\left[f_{\mathrm{VLM}}(\{I_t^{(j)}\}_{j=1}^{J},\mathcal R_v)=\mathrm{match}\right],\\
 O_t(v)&=\mathbf 1\!\left[\alpha\ell_k(v)\leq d_k^{\mathrm{odom}}\leq\beta\ell_k(v)\right],\\
 A_t(v)&=M_t(v)\land O_t(v),
\end{aligned}
 \label{eq:arrival-gate}
\end{equation}
where \(\mathbf 1[\cdot]\) is the indicator function that outputs one if the enclosed condition is true and zero otherwise, $\ell_k(v)$ is the nominal route length from the last verified anchor to $v$, $d_k^{\mathrm{odom}}$ is the odometric distance accumulated over the same interval, and $0<\alpha\leq1\leq\beta$ are configurable deployment parameters defining the admissible distance range. The gates are conjunctive rather than weighted. That is, only $A_t(v)=1$ authorizes re-anchoring and later reference promotion. Geometric proximity alone remains unverified. Identity contradictions trigger relocalization over a broader candidate set. A blockage at a confidently identified edge withholds that hop without invalidating location. Uncertain localization cannot authorize a confirmed edge update. Note that nodes encode position but not orientation. At verified node $v_k$ and acceptance time $\tau_k$,
\begin{equation}
 t_k=p_B^O(\tau_k)-R_0p_{v_k}^M
 \label{eq:translation-reanchor}
\end{equation}
updates subsequent goals in~\eqref{eq:local-point-goal} while leaving $R_0$ fixed. Heading continues to come from odometry. Because recognition is asynchronous, the correction uses odometry at acceptance rather than at image capture, and stale results are discarded if the robot has left the trigger region. Position corrections are excluded from traveled distance.

\subsubsection{Task-Sufficient Spatial Grounding}
\label{sec:goal-grounding}
\label{sec:error-isolation}
Verified anchors interrupt translational drift such that local execution need not rely on a globally accurate pose. Let $\varepsilon_{\mathrm{ver}}$ bound the planar residual between the stopped robot and the true node anchor after verification, including the node acceptance region and stopping error. Let $\varepsilon_M$ bound target-position error induced by floor-plan node coordinates, map scale, and the initial topology-odometry registration. Between correctly verified anchors, after actual travel $s\leq L_k$, denote the upper bound of planar position error by $\varepsilon_{\mathrm{ver}}+\rho s$, the upper bound of map-position error by $\varepsilon_M$, the upper bound of heading error by $\bar\theta\leq\pi$, and the upper bound of true local-goal distance by $D_{\max}$. Using Equation~\eqref{eq:local-point-goal}, comparing $g_t^B$ with the true robot-frame goal $\bar g_t^B$ one has
\begin{equation}
 \|g_t^B-\bar g_t^B\|\leq
 \varepsilon_{\mathrm{ver}}+\rho L_k+\varepsilon_M
 +2D_{\max}\sin(\bar\theta/2).
 \label{eq:error-bound}
\end{equation}
The preceding relations are operationally coupled. The point goal in~\eqref{eq:local-point-goal} uses the latest translational correction, which is updated by~\eqref{eq:translation-reanchor} only when the arrival gate in~\eqref{eq:arrival-gate} accepts a semantic node. Let $\Delta_{\pi,c}$ denote the empirical local-goal perturbation tolerance of local policy $\pi$ under scene condition $c$. A conditional compatibility requirement for verification interval $k$ is
\begin{equation}
 B_k = \varepsilon_{\mathrm{ver}}+\rho L_k+\varepsilon_M
 +2D_{\max}\sin(\bar\theta/2) \leq \Delta_{\pi,c}.
 \label{eq:policy-compatibility}
\end{equation}
Equation~\eqref{eq:error-bound} supplies the bound $B_k$ under the stated assumptions. Neither it nor~\eqref{eq:policy-compatibility} guarantees navigation success. Motivated by capability matching in feedback roadmaps and funnel composition~\cite{agha2014firm,majumdar2017funnels}, the condition makes explicit the interface between accumulated perturbation and local-policy capability. AVALON limits the translational contribution $\rho L_k$ by starting a new interval after each correctly verified semantic anchor. Geometric waypoints do not delimit this interval. The current system neither estimates $\Delta_{\pi,c}$ online nor enforces~\eqref{eq:policy-compatibility} during skeleton construction. The tolerance must be calibrated empirically for each policy and scene. Position-only re-anchoring does not remove yaw or map error. From the above, one can formulate the tackled conditional feasibility problem as follows.
\begin{equation}
\begin{aligned}
\textbf{Find}\quad & v_k,\tau_k,t_k \\
\mathrm{s.t.}\quad
&A_t(v_k) = M_t(v_k)\land O_t(v_k) = 1, \\
&t_k = p_B^O(\tau_k)-R_0 p_{v_k}^M, \\
&B_k = \varepsilon_{\mathrm{ver}}+\rho L_k+\varepsilon_M +2D_{\max}\sin(\bar\theta/2) \le \Delta_{\pi,c}.
\end{aligned}
\label{eq:feasibility-program}
\end{equation}

After the arrival at goal-associated topological nodes is verified by our proposed technique, an onboard open-vocabulary detector in~\cite{cheng2024yoloworld} will be applied to analyze the real-time observations to localize the target object. The object-search mission completes exclusively upon successful target detection, as a single negative observation is insufficient to assert object absence and necessitates further perceptual sampling. Notably, this semantic confirmation operates without precise global pose estimates. 

\subsection{Task-Triggered, Evidence-Gated Human-Robot Memory Co-Updating}
\label{sec:co-maintenance}
AVALON links memory maintenance to execution through verified graph IDs. Human and robot inputs exhibit asymmetric authority. Specifically, user configure permissions and log environment changes beyond robot observations, whereas the robot supplies localized physical evidence. A pending update becomes a sensing objective when its corresponding node or edge affects the current route, allowing execution to selectively refresh memory for subsequent missions. During a mission, the operational model $\mathcal M_t=(V,E,\Phi_t)$ updates only the state features $\Phi_t$, while the deployed topology  $(V,E)$ remains fixed. Authorized users may revise the graph before a subsequent deployment.


For each graph element $x\in V\cup E$, its task memory separates three fields:
\begin{equation}
 \phi_t(x)=\bigl(s_t(x),\ell_t(x),p_t(x)\bigr),
 \label{eq:decoupled-fields}
\end{equation}
where $s_t(x)\in\{\text{unknown},\text{clear},\text{blocked}\}$ is the robot-observable physical condition, $\ell_t(x)$ is an open-vocabulary semantic description, and $p_t(x)\in\{\text{allowed},\text{forbidden}\}$ is a human-authorized task permission. These fields are decoupled by update authority, not assumed statistically independent. Human reports remain pending hypotheses over affected fields rather than accepted physical state.

\subsubsection{Complementary authority}
Localized robot evidence may revise $s_t(x)$ and attach an observed $\ell_t(x)$, but cannot modify $p_t(x)$. Authorized users set permissions, but permission does not establish physical clearance. Updates are field-specific, i.e., a confident clearance assessment may revise $s_t(x)$ while the obstacle label remains uncertain, whereas a semantic correction does not itself establish traversability. A robot-observed blockage always stops execution.

\subsubsection{Task-relevant verification}
If the affected node or edge is required by the current route, the robot checks it during that mission. Otherwise, the report remains pending for a later relevant task. Onboard open-vocabulary detections identify object-related conditions, while depth and local feasibility checks assess immediate geometric clearance. These checks update memory evidence but never replace the online safety controller.

\subsubsection{Evidence-gated reuse}
A physical-state update requires a resolved graph ID and localized robot evidence. A doorway observation may update its entrance-node condition, whereas edge traversability additionally requires both endpoints and evidence covering that hop. Node-reference promotion still requires $A_t(v)=1$ from Section~\ref{sec:execution}. Records retain source, time, and evidence. Stale or conflicting revisions remain pending. Memory identifies where evidence is needed, and verified execution determines which observations can revise it without replacing online safety checks.

\section{Experiments}
\label{sec:experiments}
\subsection{Experimental Setup}
\begin{figure*}[!t]
\centering
\includegraphics[width=\textwidth]{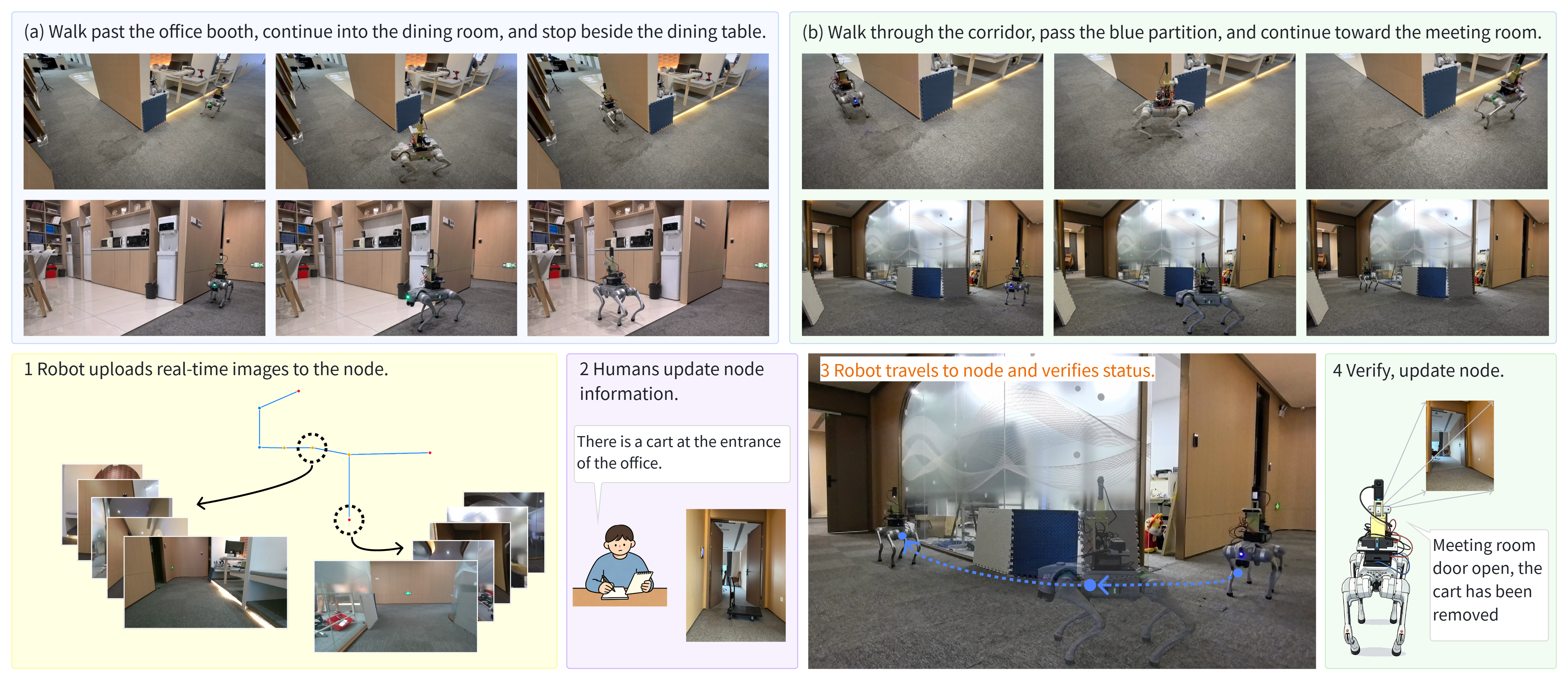}
\caption{Qualitative navigation and memory co-updating. (a)~Intersection-to-Dining first-mission route. (b)~Dining-to-Meeting-Room co-updating case: a user reports a cart at the office entrance, the robot checks the location, records the entrance node as clear, and reaches the meeting room.}
\label{fig:realworld}
\end{figure*}

Experiments use a Unitree Go2 with an Insta360 X3 $360^\circ$ camera, an Intel RealSense D455F RGB-D camera, and an NVIDIA Jetson AGX Orin. NavDP and X-NavDP send RGB-D observations over ROS~2 to an RTX~5090 workstation, which returns local waypoints at 6~Hz. Semantic-node verification calls Qwen3.5-Omni~\cite{qwen2026qwen35omni} through its API. YOLO-World~\cite{cheng2024yoloworld} runs on the Jetson at 15~Hz for object-related memory verification and endpoint target confirmation (a red trash bin in Dining or a chair in the Meeting Room). Odometry uses the Unitree \texttt{/odom} interface. The office scene has an 8-node floor-plan graph (24.11~m central trunk) seeded with 5 user photos. Timed routes are Intersection-to-Dining (12.81~m, 6 nodes) and Dining-to-Meeting-Room (20.43~m, 7 nodes). The five AVALON + X-NavDP Dining-to-Meeting-Room trials in Table~\ref{tab:main-results} also evaluate memory co-updating; their 78.7~s includes cart-report verification and memory update (Fig.~\ref{fig:realworld}(b)).
Each method is evaluated in five trials from a comparable start pose. A trial is marked successful if the robot reaches the designated goal region. AVALON requires no on-site robot mapping pass and topology--odometry alignment is initialized once at deployment time.

\begin{figure}[!t]
\centering
\includegraphics[width=\columnwidth]{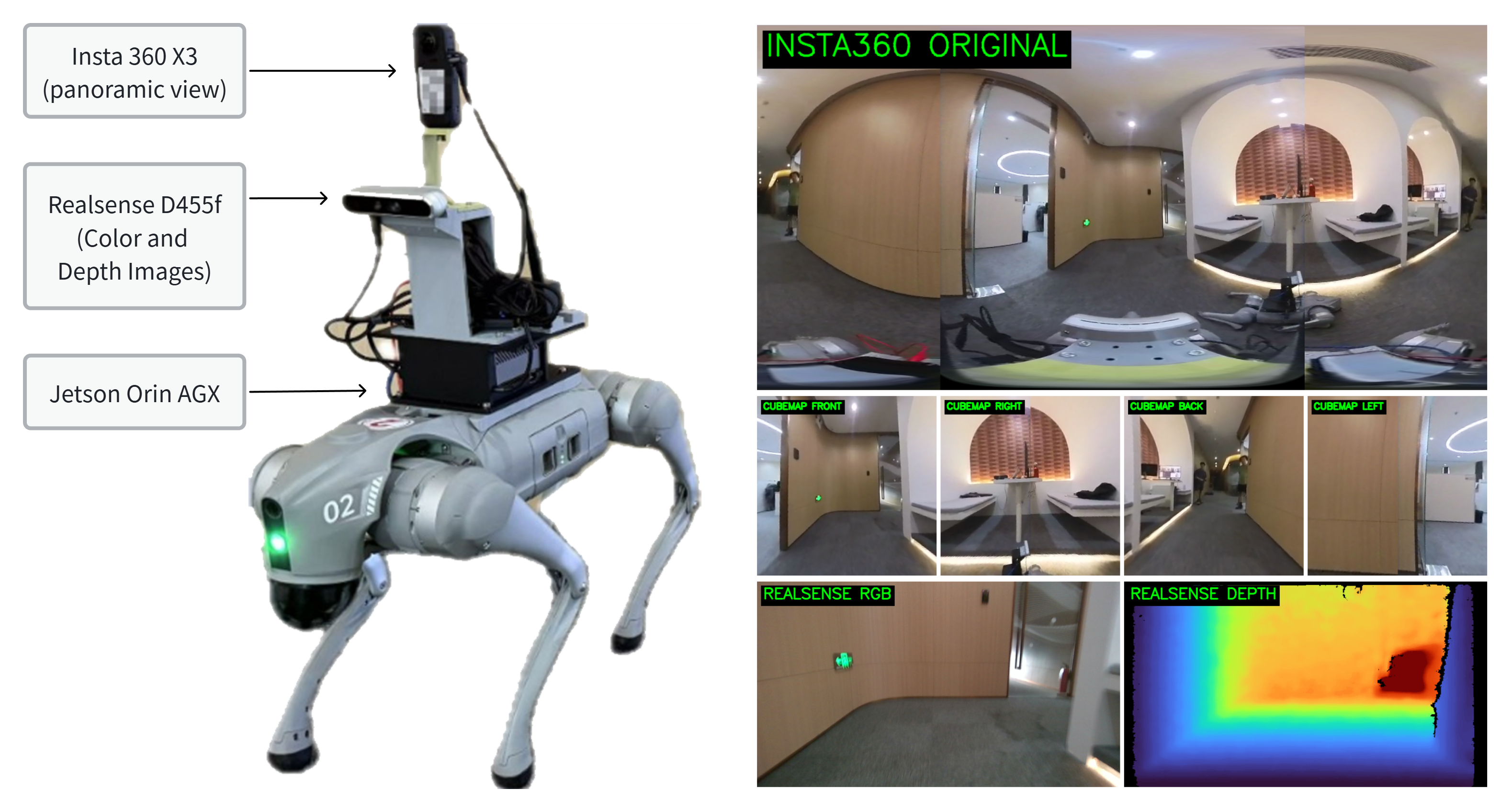}
\caption{Left: Go2 sensing platform. Right: equirectangular view, four cubemap projections, and RGB-D observations for node verification and local execution.}
\label{fig:robot}
\end{figure}

\textbf{Baselines.} \emph{DimOS-Go2}, based on the DimOS agentive operating system~\cite{dimensionalos2026dimos}, maps a region slightly larger than the task workspace (onboard odometry, WebRTC LiDAR, VoxelBlockGrid with column carving, height/slope costmap, replanning A*, no loop closure) at 0.55~m/s, then navigates. Policy-only NavDP~\cite{cai2025navdp} and X-NavDP~\cite{yang2026xnavdp} receive one route-destination point goal without the AVALON graph, node verification, or re-anchoring. We pair each with its AVALON-enabled counterpart under the same route, 0.50~m/s speed limit, and 120~s timeout.

\textbf{Metrics.} We report success rate, mean execution time over successful trials, and robot deployment time $T_{\mathrm{dep}}=T_{\mathrm{mapping}}+T_{\mathrm{execution}}$. For all methods, success requires reaching the goal region and detecting its target with YOLO-World within 120~s. If all trials fail, execution and deployment equal the timeout. Human-prior production, map I/O, and pose initialization are excluded from $T_{\mathrm{dep}}$. 


\begin{table}[htp]
\centering
\caption{First-mission results on two routes in the same office (five trials per method and route), where ``policy-only'' removes the AVALON graph, verification, and re-anchoring.}
\label{tab:main-results}
\scriptsize
\setlength{\tabcolsep}{2.3pt}
\begin{tabular}{@{}lcccc@{}}
\toprule
Method & Success rate & Mapping (s) & Execution (s) & Deployment (s) \\
\midrule
\multicolumn{5}{@{}l}{\emph{Intersection $\rightarrow$ Dining (12.81~m)}}\\
DimOS-Go2 & $4/5$ (80\%) & 45.2 & 33.3 & 78.5\\
NavDP (policy-only) & $0/5$ (0\%) & 0 & 120 & 120 \\
\textbf{AVALON + NavDP} & $3/5$ (60\%) & 0 & 83.4 & 83.4 \\
X-NavDP (policy-only) & $0/5$ (0\%) & 0 & 120 & 120 \\
\textbf{AVALON + X-NavDP} & $5/5$ (100\%) & 0 & 46.4 & \textbf{46.4} \\
\midrule
\multicolumn{5}{@{}l}{\emph{Dining $\rightarrow$ Meeting Room (20.43~m)}}\\
DimOS-Go2 & $5/5$ (100\%) & 60.2 & 52.3 & 112.5 \\
NavDP (policy-only) & $0/5$ (0\%) & 0 & 120 & 120 \\
\textbf{AVALON + NavDP} & $1/5$ (20\%) & 0 & 93.6 & 93.6 \\
X-NavDP (policy-only) & $0/5$ (0\%) & 0 & 120 & 120 \\
\textbf{AVALON + X-NavDP} & $5/5$ (100\%) & 0 & 78.7 & \textbf{78.7} \\
\bottomrule
\end{tabular}
\end{table}

\subsection{Room-to-Room Deployment}
Table~\ref{tab:main-results} compares first-mission robot time and pairs each local policy with and without AVALON. With the X-NavDP backend, AVALON completes the Intersection-to-Dining mission in 46.4~s versus 78.5~s for DimOS, 40.9\% less, because it skips on-site mapping. Per-route execution is slower (46.4~s vs.\ 33.3~s): AVALON pauses at semantic nodes to verify identity, and the 0.05~m/s speed-limit gap alone does not explain the difference. The unsuccessful DimOS trial drifted in odometry and entered an unintended dead-end corridor. On the 20.43~m Dining-to-Meeting-Room route, AVALON with X-NavDP achieves $5/5$ success in 78.7~s deployment time, compared with 112.5~s and $5/5$ success for DimOS-Go2. Policy-only X-NavDP receives one route-destination point goal; all five trials on each route reach the 120~s timeout. Figure~\ref{fig:realworld}(a) shows the Intersection-to-Dining route.

\subsection{Robustness to Odometry Perturbations}
\label{sec:ablation}
Table~\ref{tab:main-results} establishes the benefit of the graph-guided interface. Table~\ref{tab:ablation} assesses the sensitivity of full AVALON to odometry perturbations while keeping the graph, route, X-NavDP backend, semantic-node verification, $0.50$~m/s speed limit, and 120~s timeout fixed. The perturbation model adds zero-mean Gaussian increment noise and a random-walk bias before accumulating the noisy pose. Its dimensionless parameter \texttt{odom\_noise\_strength} controls the noise scale and is set to 0.02 and 0.04 for the low and high conditions, respectively. This experiment evaluates robustness to increasing noise rather than isolating the effect of re-anchoring.

\begin{table}[htp]
\centering
\caption{Sensitivity of full AVALON to controlled odometry noise. All settings use the same graph, 12.81~m route, and X-NavDP local policy (five trials). Execution time averages successful trials only.}
\label{tab:ablation}
\scriptsize
\setlength{\tabcolsep}{5.0pt}
\begin{tabular}{@{}lccc@{}}
\toprule
Odometry setting & \texttt{odom\_noise\_strength} & Success rate & Execution (s) \\
\midrule
Nominal & 0 & $5/5$ (100\%) & 46.4 \\
Perturbed (low) & 0.02 & $3/5$ (60\%) & 44.1 \\
Perturbed (high) & 0.04 & $1/5$ (20\%) & 60.1 \\
\bottomrule
\end{tabular}
\end{table}

\subsection{Memory Co-Updating Case Study}
\label{sec:maintenance-evaluation}
In this experiment, a node ID is specified, while an ambiguous edge requires the other endpoint. Across five Dining-to-Meeting-Room deployments (Fig.~\ref{fig:realworld}(b)), a user reports a cart at the office entrance. In each trial, the robot travels to that node, observes that the doorway is clear, records the entrance-node condition, and reaches the meeting room. The report triggers verification rather than being treated as confirmed occupancy. This node-level observation does not certify an unobserved edge. These five deployments illustrate how a reported change is verified before it updates memory. 

\section{Conclusion}
\label{sec:conclusion}

This paper proposes AVALON, a topology-first indoor navigation framework that decouples task execution from rigid global map alignment.
A floor-plan-derived topological skeleton paired with editable task memory enables hierarchical mission execution. Topological routing outputs robot-frame point goals for a local navigation policy, and verified node arrivals update translational alignment without requiring yaw correction. For object-search tasks, terminal perceptual validation confirms target presence in contrast to treating node arrival as task completion. Verified robot observations feed into reference-memory updates and enable auditable memory refinement. On a Unitree Go2 first-mission deployment (12.81~m Intersection-to-Dining, workspace trunk 24.11~m, five trials), AVALON with X-NavDP succeeds in all five runs in 46.4~s without robot mapping, 40.9\% less robot time than DimOS-Go2 (45.2~s mapping + 33.3~s mean execution, $4/5$ success). Policy-only X-NavDP fails in all five trials under a 120~s timeout, pacing near the start without a long-horizon route. DimOS's unsuccessful trial drifted in odometry into a dead end. In the five AVALON + X-NavDP Dining-to-Meeting-Room trials reported in Table~\ref{tab:main-results}, user reports of environmental changes trigger on-site robot verification before the entrance node condition is updated. AVALON successfully completes the report--verification--update sequence and subsequently reaches the meeting room in all five trials. These demonstrate the superiority of the proposed topological navigation framework.




\bibliographystyle{IEEEtran}
\bibliography{references/bib/references}

\end{document}